\documentclass[letterpaper, 10 pt, conference]{ieeeconf}  

\IEEEoverridecommandlockouts                              

\usepackage{graphicx} 
\usepackage{amsmath} 
\usepackage{algorithm}
\usepackage{algpseudocode}

\title{\LARGE \bf
Estimating Material Properties of Interacting Objects Using Sum-GP-UCB
}

\author{M. Yunus Seker$^{1}$ and Oliver Kroemer$^{1}$
\thanks{*This work is supported by Sony AI and NSF Grants No. CMMI-1925130 and IIS-1956163.}
\thanks{$^{1}$The Robotics Institute, School of Computer Science,
        Carnegie Mellon University, 5000 Forbes Avenue, Pittsburgh, PA, USA
        {\tt\small mseker@andrew.cmu.edu, okroemer@andrew.cmu.edu}}%
}

\begin{document}

\maketitle
\thispagestyle{empty}
\pagestyle{empty}

\begin{abstract}

 Robots need to estimate the material and dynamic properties of objects from observations in order to simulate them accurately. We present a Bayesian optimization approach to identifying the material property parameters of objects based on a set of observations. Our focus is on estimating these properties based on observations of scenes with different sets of interacting objects. We propose an approach that exploits the structure of the reward function by modeling the reward for each observation separately and using only the parameters of the objects in that scene as inputs. The resulting lower-dimensional models generalize better over the parameter space, which in turn results in a faster optimization. To speed up the optimization process further, and reduce the number of simulation runs needed to find good parameter values, we also propose partial evaluations of the reward function, wherein the selected parameters are only evaluated on a subset of real world evaluations. The approach was successfully evaluated on a set of scenes with a wide range of object interactions, and we showed that our method can effectively perform incremental learning without resetting the rewards of the gathered observations.
\end{abstract}

\section{Introduction}

Physics simulators have become ubiquitous in robotics as they serve a number of critical roles in both planning and learning. 
However, the utility of a simulator depends largely on its ability to accurately predict the effects of actions in the real world. \emph{System identification (SysID)} methods are therefore often used to estimate the physical properties of objects in the real world for accurate simulation \cite{sysiddynamics, sysidgraph}. As the capabilities of simulators increase, and they are able to model more complex and deformable objects \cite{Herguedas2019SurveyOM, deformable, ArriolaRios2020ModelingOD}, so does the need for more accurate SysID approaches. 

Modeling objects for manipulation can generally be divided into identifying two types of parameters: geometric parameters and material/dynamic parameters \cite{li2020learning, driess2022learning}. Geometric object parameters, such as its 3D mesh, can often be acquired from static scenes using vision-based approaches. For dynamic parameters, the robot needs to observe the objects interacting and the changes (or lack thereof) resulting from these interactions. Although we present a basic pipeline for estimating geometric parameters from vision, the focus of this paper is on the SysID of the material and dynamics parameters.

Current methods often approach the SysID problem in a highly controlled manner. Objects are often presented to the robot in isolation, such that it can focus on its specific properties.  Some methods allow the robot to apply exploratory actions, such as poking objects, to estimate the material properties \cite{densephysnet}. In practice, robots should avoid excessive exploration and attempt to gather information from any interactions. 
Objects will often also not be observed in isolation but rather interacting with other (unfamiliar) objects in the scene. The sets of objects in each scene will also vary, with some objects being shared and others not. 

Similar to other recent approaches to SysID, we use Bayesian optimization (BO) \cite{bo} to determine the latent material parameters of objects. However, we focus on the case where scenes include different sets of interacting objects. The BO process involves selecting sets of parameter values, evaluating these values in simulation, comparing the resulting predictions to the observed real-world effects to obtain the reward value, and updating the BO's model with this reward information to select the next set of parameters. A key insight of our method is that the reward function being optimized consists of a sum of reward functions over individual real-world observations, and each of these reward functions only depends on the parameters of the objects in the corresponding scene. Hence, we propose to model the individual observations' reward functions using lower-dimensional Gaussian processes, rather than using one high-dimensional Gaussian process for the total reward. The resulting reward model exploits the inherent structure of the reward function and provides improved generalization over the parameter space, which in turn leads to a faster optimization.  

\begin{figure*}
    \centering
    \includegraphics[width=0.8\linewidth]{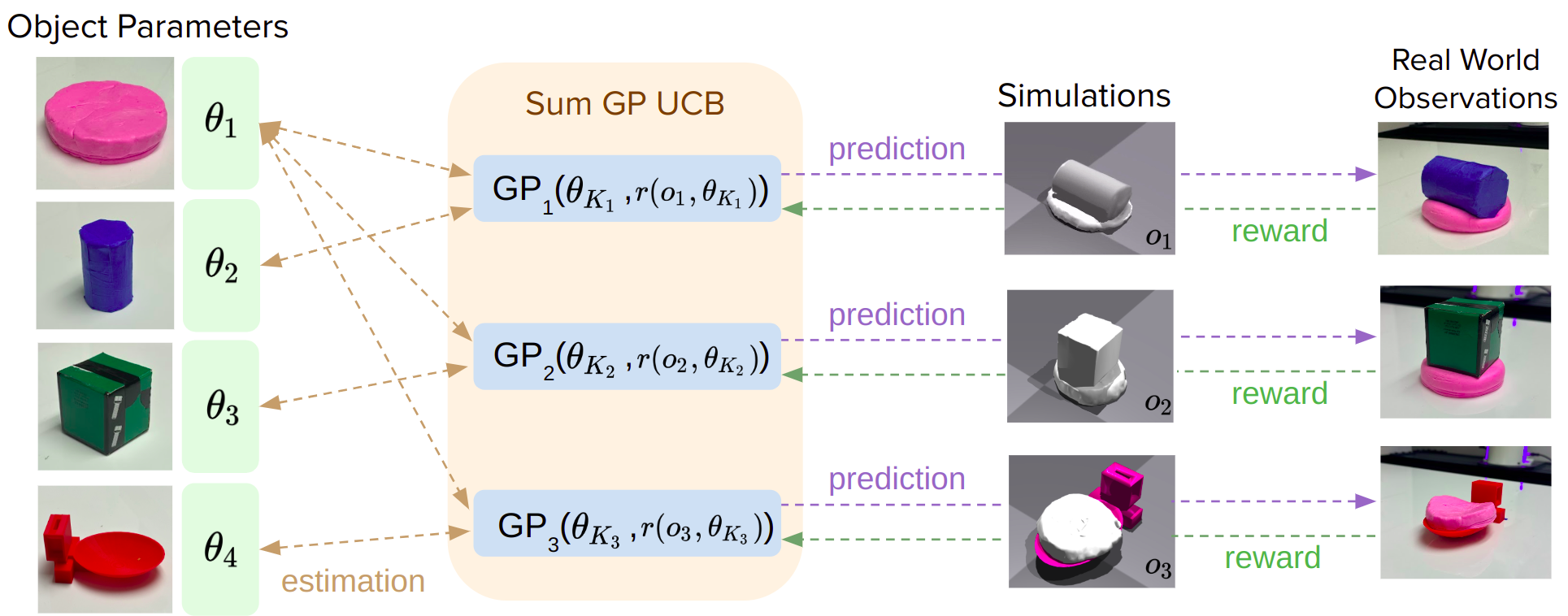}
    \caption{Overview of our proposed framework. At its core, our framework processes a collection of objects and real-world observations using simulations. The primary objective of this study is to identify the optimal set of material parameters for these objects to best reproduce the dynamics of the real world. }
    \label{fig:overview}\vspace{-4mm}
\end{figure*}

Simulation runs are a core step of the optimization process. However, these simulations can be time consuming and computationally expensive. Although our proposed reward model already reduces the number of simulation runs required to find a suitable set of material properties, we can reduce the number of runs further. In particular, we propose \emph{partial} reward evaluations.  In this case, only a subset of the observation reward functions are evaluated. Such partial evaluations are made possible by modeling the observation rewards individually. We propose an upper confidence bound (UCB)  BO strategy for selecting the next set of object parameters to evaluate and an exploration strategy for selecting the observation rewards to evaluate with these parameters. We refer to the resulting approach as \textbf{Sum GP UCB}. An overview of the approach is shown in Fig. \ref{fig:overview}. 

By modeling individual observation rewards and allowing for partial evaluations, the Sum GP UCB approach presents an additional benefit: allowing for incremental observations. If the robot is performing SysID online, it may continue to acquire additional observations while the identification process is running. In theory, this represents a change in the overall reward function, including potentially the space of input parameters. However, for Sum GP UCB it simply requires an additional low-dimensional reward model for the new observation. 
In summary, our key contributions are: 
\begin{enumerate}
    \item  We model the reward function for each observation individually using only the relevant object parameters.
    \item We use partial evaluations to further reduce the number of simulation runs needed by the optimization process. 
    \item We demonstrate our method's ability to handle adding observations during the identification process. 
\end{enumerate}
We successfully evaluated the proposed framework using a set of scenes with pairs of interacting objects. The experiments show that the benefits of the proposed framework increase with the number of observations being optimized over. The experiments also show that our approach with partial evaluations even outperforms the standard BO approach with all observations evaluated at each iteration.

\vspace{-3pt}
\section{Related Work}
In recent years, differentiable simulators \cite{drake} have become a suitable tool for system identification for task optimization and material estimation, such as estimating friction coefficient \cite{friction} and learning to slide objects \cite{slide}. Using Finite-Element-Method (FEM) \cite{hughes2012finite} and Position-Based-Dynamics (PBD) \cite{pbd} various methods have been proposed introducing soft object tasks including robot planning and control \cite{hu2018chainqueen}, grasp manipulation \cite{huang2021defgraspsim}, tactile sensing \cite{narang2021simtoreal}, optimal motion planning \cite{7428219}, and reinforcement learning \cite{matas2018simtoreal}. Custom differentiable simulators allowed researchers to train end-to-end systems to identify the materials of the objects observed in the real-world. As one of these studies, GradSim \cite{murthy2021gradsim} uses a custom differentiable simulator to estimate the material properties of the objects from real-world videos by back-propagating image pixel errors. However, as the most of the studies in this area, their study also focuses on single objects with interactions mostly covered by gravity and ground collision.

Combined with the simulations, Bayesian optimization is effectively used in the field of robotics for various applications such as domain randomization for policy searching \cite{ramos2019bayessim}, parameter estimation  \cite{antonova2021bayessimig}, and robotic cutting \cite{heiden2021disect}.  Among these studies, the work closest to ours \cite{pmlr-v205-antonova23a}, combines global Bayesian optimization with local leaps to achieve action parameters and material properties estimation for task optimization and system identification. Their study, while closely related to ours, primarily focuses on single-task optimization with few object interactions. In contrast, our approach takes advantage of multi-object multi-interaction settings where the material dynamics of the objects emerge more qualitatively due to the diversity of the environments.

\section{Technical Approach}

\subsection{Problem Formulation}
The goal of the proposed work is to optimize the material parameters $\theta=\{\theta_1,...\theta_m\}$ of a set of $m$ objects in simulation based on a set of $n$ real world observations  $o_1,...,o_n$. Each observation $o_i$ includes an initial scene $s_{i0}$, and action $a_i$, and a final scene $s_{iT}$. Each observation is associated to a set of objects $K_i\subset{1,...m}$ that are present in the scene. 

The robot is provided with a simulation engine $\hat{s}_T=f(s_0,a,\theta)$ with which it can predict the final state $s_T$ given the object parameters $\theta$, the initial state $s_0$, and the action parameters $a$. As only a subset of the objects is present in each scene, only the corresponding object parameters will have an effect such that $f(s_{i0},a,\theta)=f(s_{i0},a,\theta_{K_i})$.

The robot should optimize the object parameters to minimize the error between the predicted and real final states. We assume that the reward function has the form $R(o,\theta)=\sum_{i=1}^n r(o_i,\theta_{K_i})$. Evaluating the observation reward functions $r(o_i,\theta_{K_i})$ involves running the simulation for observation $o_i$ with a set of material parameters $\theta_{K_i}$. The goal is to find an optimal set of material parameters $\theta^*=\arg\max_\theta R(o,\theta)$.

\begin{figure}[b]
    \centering
    \includegraphics[width=\linewidth]{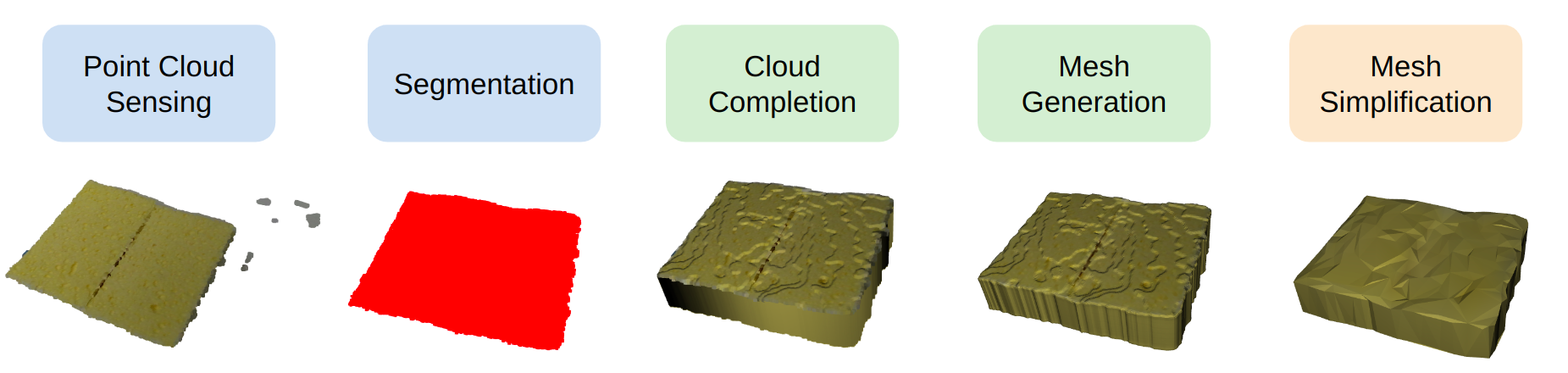}
    \caption{From left to right the figure shows the key steps of our mesh generation process.}
    \label{fig:meshgen}
\end{figure}

\subsection{Geometric Real-to-Sim Scene Processing}
Although the focus of the paper is on estimating the latent material properties of the objects being manipulated $\theta$, the robot's simulator $f$ also needs to take into consideration the observed geometry of the objects. Rather than relying on handcrafted models of the objects, we developed a pipeline for generating 3D object models, that can be simulated, based on observed tabletop environments. Our approach assumes that the object is resting on a supporting surface and has a simple shape. It should be noted that other 3D model estimation methods \cite{wen2023bundlesdf} could also be used within our proposed framework. 

The mesh generation pipeline is shown in Fig. \ref{fig:meshgen}. The pipeline takes a partial top-down point cloud of the object of interest, as well as the table's normal direction as input. Spurious clusters are segmented out of the cloud using the DBSCAN \cite{dbscan} algorithm. The shape is completed by extruding points into the table surface to create a wide base.  The extrusion is performed by first detecting the edge points in the partial point cloud using the concave hull algorithm. These points are incrementally repeated into the table plane to fill in hidden surfaces on the sides of the object. To create the bottom support surface, all of the points in the mesh are projected into the table plane. Given the extrusion-completed point cloud, we generate a mesh using the Poisson surface reconstruction algorithm \cite{surfaceReconstruction} with parameters $(depth=8, scale=2.0)$. To make the mesh easier to simulate, the final step is to decimate its polygon count down to 500.

This extrusion-based approach is not suitable for all classes of objects, but it works well for deformable objects (such as dough or mashed potatoes) and simple shapes (such as boxes and cylinders) \cite{lin2022diffskill, dough}. One could also augment the environment with additional cameras, but our experiments show that the robot can achieve good performance even with this simple approach. The near top-down images are natural for tabletop manipulation tasks and help to ensure that we acquire accurate models for the heightmap-based reward functions, as described in Section III.D. 

\begin{figure*}
    \centering
    \includegraphics[width=0.9\linewidth]{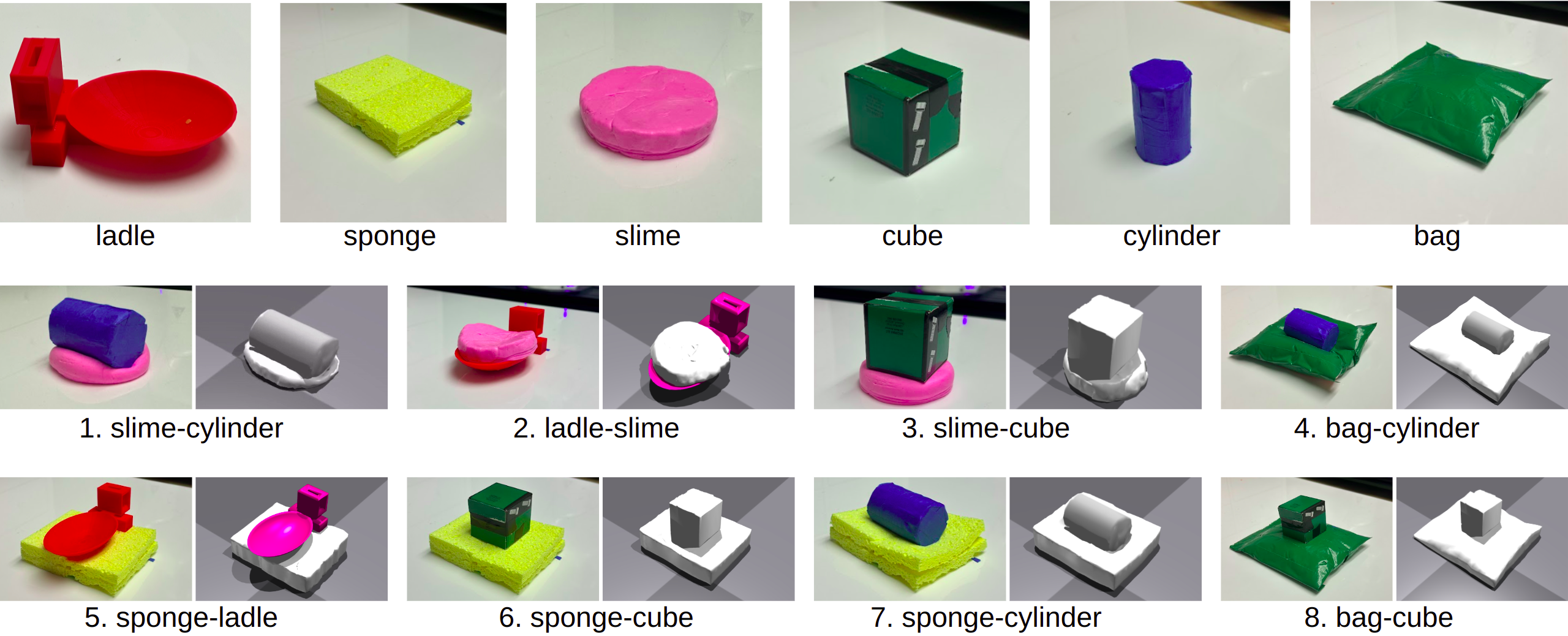}
    \caption{Our dataset and the scenes used in the experiments. (Top) The dataset includes a wide range of deformable (sponge, slime, bag) and rigid objects (ladle, cube, cylinder) with different material properties. (Bottom) Real-world scenes and their corresponding simulations side by side.}
    \label{fig:sim_real_all}\vspace{-4mm}
\end{figure*}

\subsection{Dynamics Simulation}
Given the 3D object meshes of the initial state $s_i$, a set of object parameters $\theta$, and the action parameters $a$, the robot can use its simulation engine to predict the resulting state $s_T$. For our evaluations, we have used simple pick-and-place actions to create interactions between pairs of objects. The action parameters define the location where the centroid of the top object is placed relative to the bottom object. In the real world, the robot arm is used to grasp the top object and place it at the desired location. The finger closure is set to avoid large deformations of the object. The simulation of the grasping interactions has a tendency to become unstable. We therefore simulate the action by directly shifting the position of the top object's mesh to slightly above the desired location (Fig \ref{fig:sim_real_all}). To simulate the interactions and objects, we use NVIDIA's Isaac Gym \cite{isaacgym}. The Flex engine plugin is used for modeling deformable objects. It should be noted that the current version of the simulator does not handle soft-soft object contacts. Our data therefore focuses on pairs of objects, although the framework can in theory handle multiple objects in each scene.

The set of material parameters depends on the type of objects. For rigid bodies, the robot needs to estimate the mass of the object. For deformable objects, the robot needs to estimate the Young's modulus and the Poisson's ratio. 
For the purpose of system identification, we assume that the object parameters are within the ranges of $[0.01, 2]$ kg for masses, $[1000, 10000]$ for Young's modulus, and $[-0.5, 0.5]$ for Poisson's ratio.

\begin{figure}[b]
    \centering
    \includegraphics[width=\linewidth]{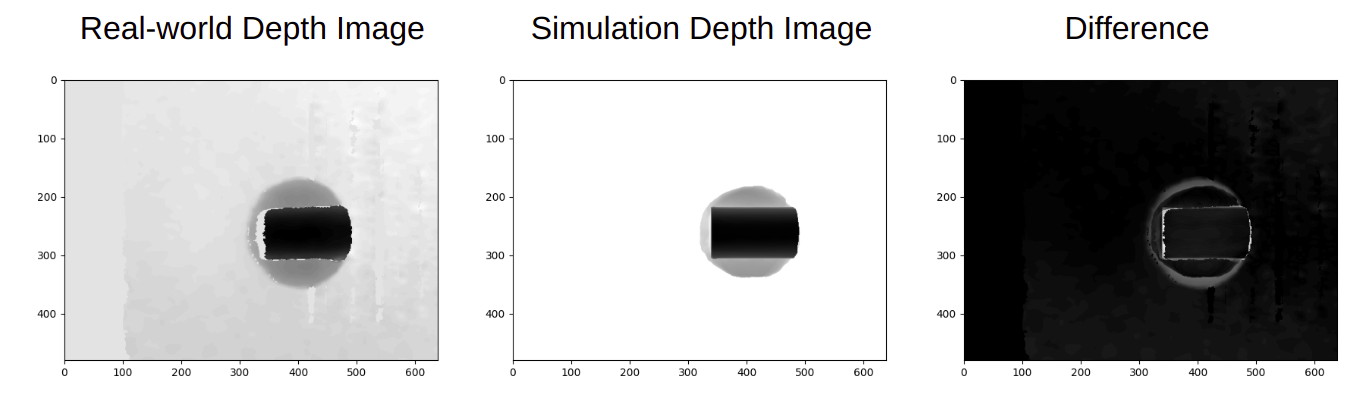}
    \caption{Example depth images from real-world and simulation. The difference between the real world and the simulation observations is visualized at the right. Darker pixels represent closer points.}
    \label{fig:sim_real_depth}
\end{figure}

\subsection{Scene Comparisons and Reward Function}
Given a ground truth final scene $s_T$ and a predicted final scene $\hat{s}_T$, the robot requires a reward function, or similarity measure, to determine the prediction accuracy. We want our reward function to be capable of capturing a variety of different simulations and object models (e.g., rigid and deformable meshes).  

We use a heightmap representation $\phi(s)$ to capture the scenes (Fig \ref{fig:sim_real_depth}). Each scene is represented by a $640 \times 480$  heightmap. The real and simulated cameras are calibrated such that their extracted heightmaps are automatically aligned. The reward function is then  defined as $r(o,\theta)=-|\phi(s_T)-\phi(f(s_i,a,\theta))|$

The heightmap is well suited for tabletop arrangement tasks as the robot will often observe the scene from above. Discrepancies lower down in the scene are not captured by this reward function, but these differences are also often difficult to observe in practice. The heightmap representation can be extracted from a variety of simulations (both learned and analytical) and thus also supports future benchmarking efforts.

\subsection{Bayesian Optimization with Sum GP UCB}
Given the ground truth observations and simulation models of the scenes, the final step is to optimize the material property parameters of the objects. We use a Bayesian optimization (BO) approach to identify a suitable set of parameters.

The standard BO approach uses a Gaussian process (GP) model to represent the target reward function $R(o,\theta)$. The GP computes a mean $\mu(\theta)$ and variance $\sigma(\theta)^2$ estimate of the reward function for any set of parameters $\theta$. The mean represents the expected reward while the variance captures the uncertainty of the reward function at that parameter setting.  To add new data to the GP, the robot must select a set of material parameter values to query $\theta$ and then evaluate these parameters by simulating each of the $n$ observations. The sum of the rewards for each observation's prediction accuracy then gives the total reward $R(o,\theta)=\sum_{i=1}^n r(o_i,\theta_{K_i})$ for that parameter setting. The next set of parameter values to evaluate $\theta'$ is selected based on an acquisition function $\alpha(\theta)$. In this paper, we use the common upper confidence bound (UCB) policy such that $\alpha(\theta)=\mu(\theta)+\beta \sigma(\theta)$, where $\beta$ is a hyperparameter that adjusts the exploration bias of the resulting policy. The next set of parameters is then given by $\theta'=\arg\max_\theta \alpha(\theta)$.  

The core insight of our proposed approach is that we can exploit the cumulative structure of the reward function $R(o,\theta)=\sum_{i=1}^n r(o_i,\theta_{K_i})$ to achieve improved generalization and partial evaluations. Summing together the individual observation reward terms  $r(o_i,\theta_{K_i})$ results in a loss of information. Instead, we propose to model each observation reward $r(o_i,\theta_{K_i})$ using a separate GP model with mean $\mu_i(\theta_{K_i})$ and variance $\sigma_i(\theta_{K_i})^2$. A key benefit of this approach is that each of the resulting GPs is defined in the lower dimensional space of $\theta_{K_i}$ rather than the full dimensional space of $\theta$. As an alternative interpretation, one could imagine each of the observation reward GPs taking in the full $\theta$ parameter, but having the kernel length scale parameters set to infinity for the object parameters that are not in $K_i$. In this manner, we can see the GPs as using kernels that are tuned for the individual observation rewards rather than shared across the total reward.

To perform the BO, we define the UCB acquisition function as :$$\alpha(\theta)=\sum_{i=1}^n\mu_i(\theta_{K_i})+\beta_i \sigma_i(\theta_{K_i})$$ The next set of parameters to evaluate is still given by $\theta'=\arg\max_\theta \alpha(\theta)$. In practice, simulation runs can be time consuming and even redundant; if one of the simulation predictions performs terribly, it may not be worth evaluating all of the other simulations with the same parameters. Fortunately, our proposed framework of tracking individual observation rewards allows us to perform \emph{partial evaluations}. In this case, the robot may only evaluate some of the simulation predictions with the selected parameters $\theta'$ and use the results to only update the corresponding GPs. To explore the potential benefits of such an approach, we restrict the robot to only selecting one simulation to evaluate per parameter setting. We select the simulation with the largest variance  $\arg\max_i \sigma_i(\theta'_{K_i})$ for the selected parameter values. This approach represents a purely exploratory strategy, and the goal is to quickly reduce the uncertainty of this optimistic parameter setting. 

The Gaussian processes were implemented using the BoTorch \cite{botorch} software library. The hyperparameters were tuned to maximize the marginal likelihood of the collected data. The next set of parameters is chosen by the built-in multi-start gradient optimizer of BoTorch. The GPs used a Matern kernel function with length scale automatically set for each GP as part of the hyperparameter tuning process.

\section{Evaluations}

\subsection{Experimental Setup}

To evaluate the proposed approach, we collected a dataset of eight observations with pairwise object interactions. The scenes consist of objects drawn from a set of six objects. The objects include a wide range of material properties for both rigid and deformable objects as shown in Fig. \ref{fig:sim_real_all}. 

For our experiments, we are evaluating our Sum-GP-UCB approach compared to the standard GP-UCB approach applied directly to the total reward $R$, which we refer to as Naive BO due to the loss of reward structure information. We are particularly interested in observing the differences in the learning curves between these two approaches. The learning curves show the total error for each selected parameter setting $-R(o,\theta')$. For our proposed approach with partial evaluations, the robot is only given the reward for the selected observation, but we still compute the total error $-R(o,\theta')$ for our learning curves. 

We present the error curves both as a function of the number of iterations and the number of simulations. The number-of-simulation results show how efficiently the approaches can utilize each simulation run. For the number-of-iteration results, the Naive BO approach computes the reward for all observations in each iteration. This result reflects the scenario where all of the simulations can be run in parallel. For our approach, we still restrict the number of simulations to one per iteration to emphasize the simulation efficiency, although one could evaluate multiple observations in parallel. 
\vspace{-4pt}
\subsection{Observation Set Size Experiment}
A distinct advantage of our approach is the capability of managing a substantial number of objects and observations due to its individual GP design. To evaluate the efficacy of our approach in different object and observation counts, we constructed three experimental sets consisting of 4, 5, and 6 objects, paired with 3, 6, and 8 observations, respectively. We compared our outcomes against the Naive Bayesian Optimization (Naive BO) approach, utilizing the metrics explained in Section IV. A. Figure \ref{fig:sim_real_all} shows the objects and the observations that we used in our experiments.

\begin{figure}
    \centering
    \textbf{\textsc{Experiment 1 - 3 Observations}\vspace{4pt}}
    \includegraphics[width=0.9\linewidth]{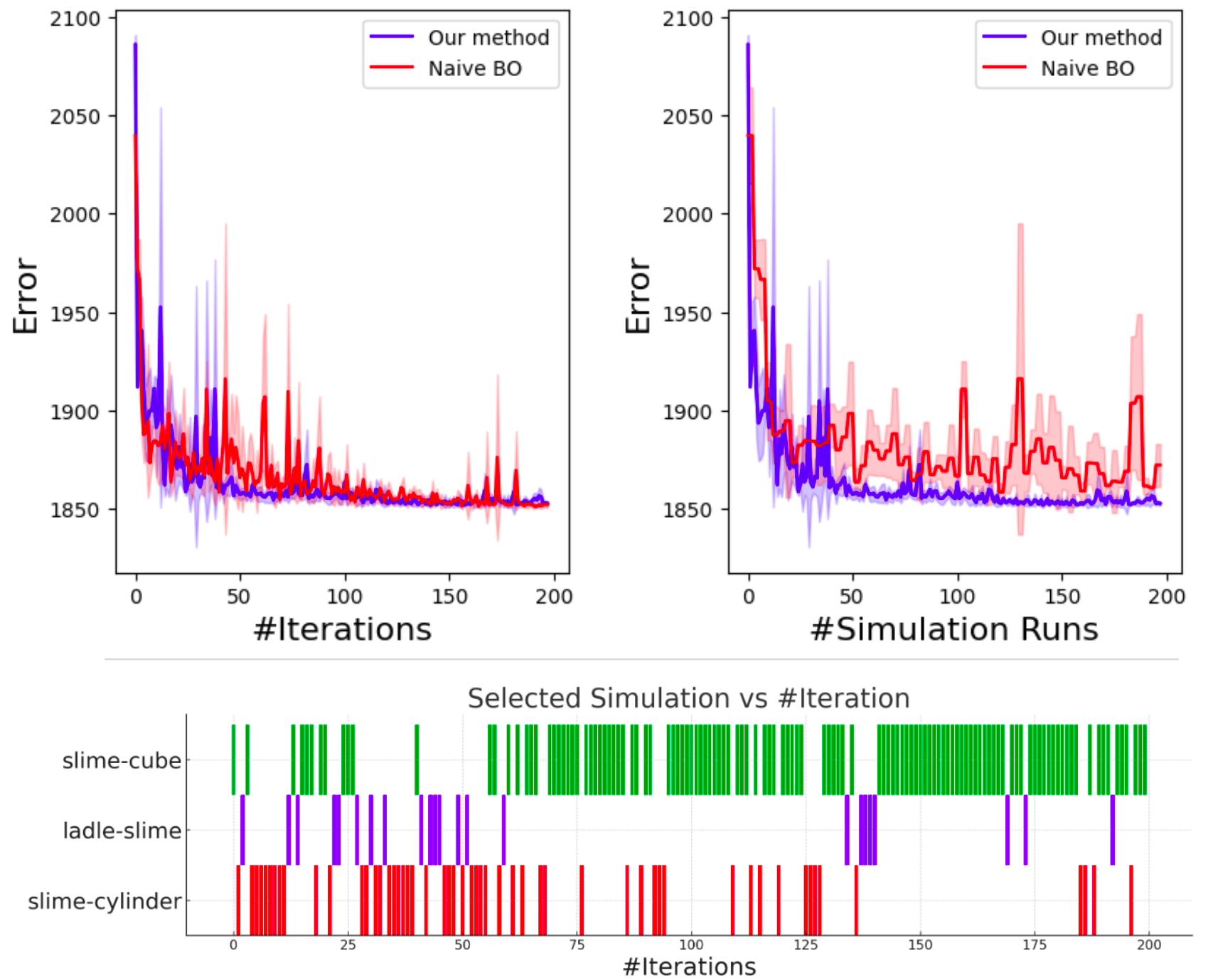}
    \vspace{-6pt}
    \caption{(Top) Learning curve plotted against the number of iterations and simulation runs. (Bottom) A sample experiment illustrating the chosen simulation for that specific iteration.}
    \label{fig:exp1} \vspace{-5mm}
\end{figure}

The first experiment set included the slime, cylinder, ladle, and cube objects as well as scenes 1, 2, and 3. We ran both Sum-GP-UCB and Naive BO methods for 200 iterations. Figure \ref{fig:exp1} shows the resulting learning curves and selected simulations at each time step for this experiment.

\begin{figure}
    \centering
    \textbf{\textsc{Experiment 2 - 6 Observations}\vspace{4pt}}
    \includegraphics[width=0.9\linewidth]{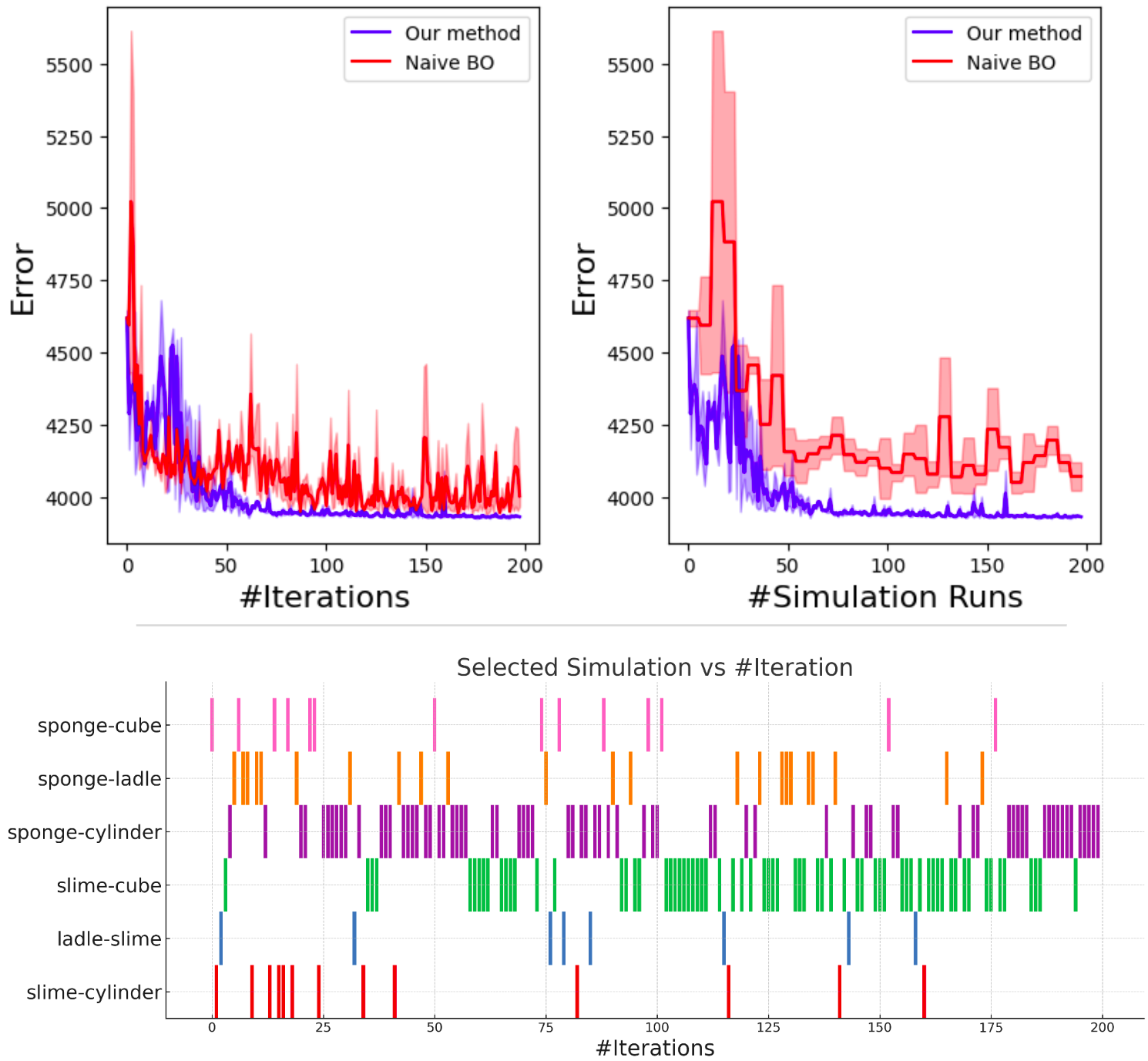}
    \vspace{-6pt}
    \caption{(Top) Learning curve plotted against the number of iterations and simulation runs. (Bottom) A sample experiment illustrating the chosen simulation for that specific iteration.}
    \label{fig:exp2}\vspace{-5mm}
\end{figure}

For our second experiment, on top of the objects and scenes in the first experiment, we also used the sponge object and included scenes 5, 6, and 7 for a total of 5 objects and 6 observations. The learning curves over 200 iterations are shown in Fig. \ref{fig:exp2}.

\begin{figure}[b]
    \vspace{-6pt}
    \centering
    \textbf{\textsc{Experiment 3 - 8 Observations}\vspace{4pt}}
    \includegraphics[width=0.9\linewidth]{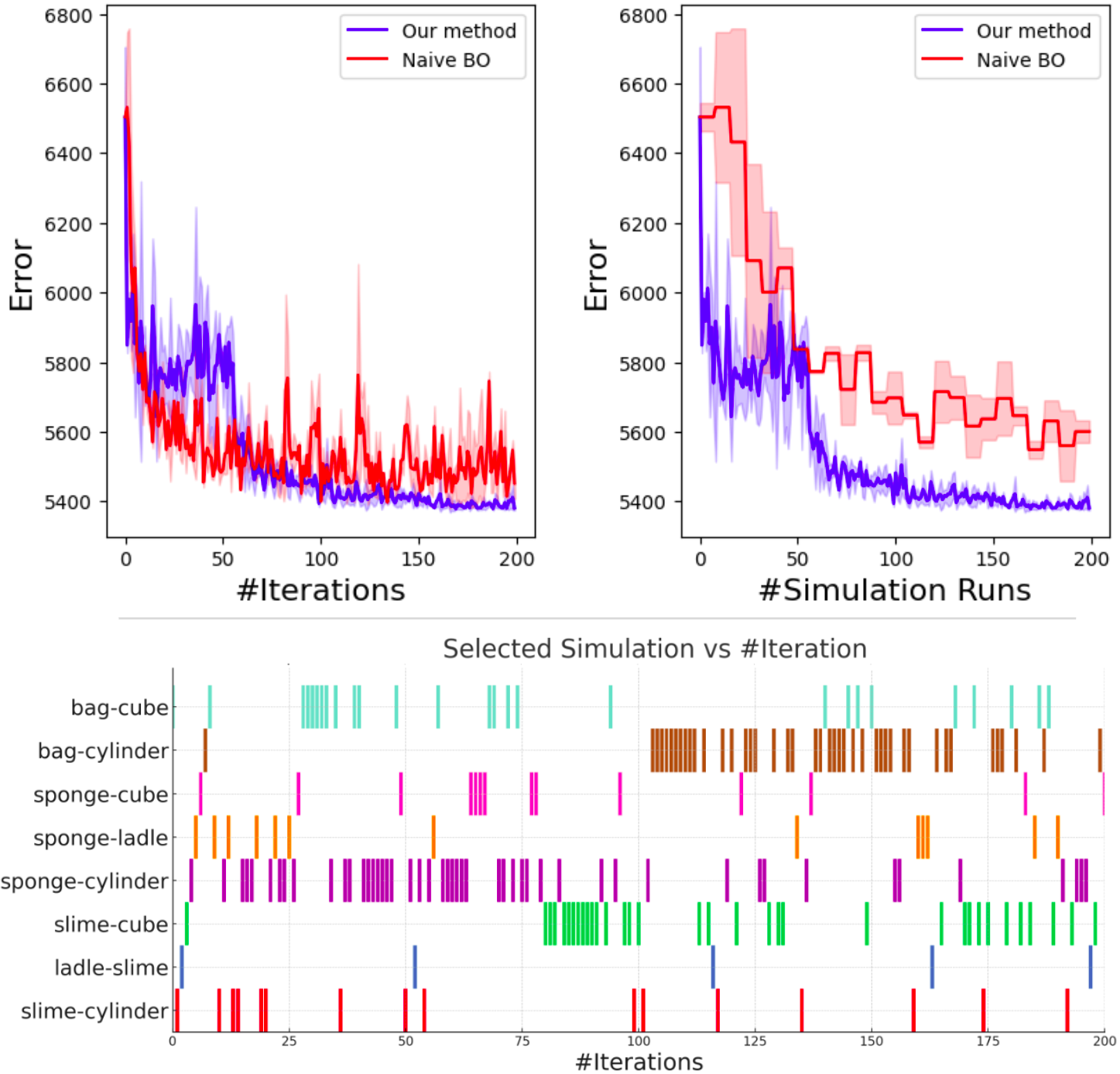}
    \vspace{-6pt}
    \caption{(Top) Learning curve plotted against the number of iterations and simulation runs. (Bottom) A sample experiment illustrating the chosen simulation for that specific iteration.}
    \label{fig:exp3}\vspace{-4mm}
\end{figure}

\begin{figure*}
    \centering
    \textbf{\textsc{Experiment 4 - Incremental Observations}\vspace{4pt}}
    \includegraphics[width=0.9\linewidth]{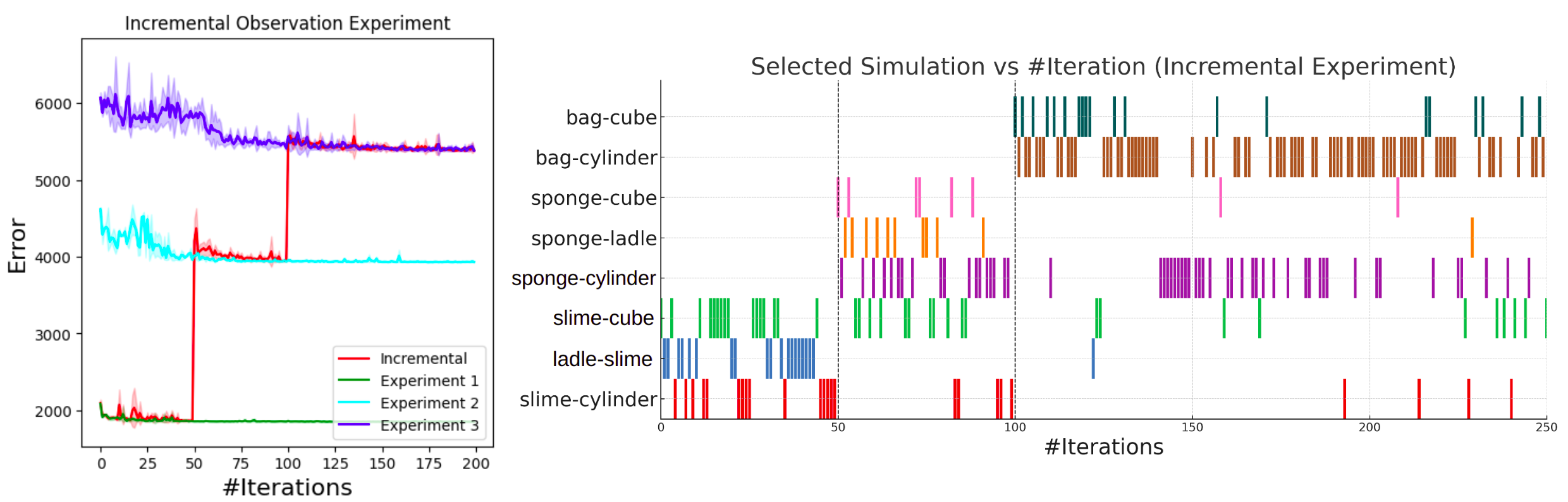}\vspace{-3mm}
    \caption{The new observations are introduced at iterations 50 and 100. When new observations are introduced, the algorithm focuses on learning the new scenes,it should be noted that the algorithm switches back frequently and refines its GPs for the older scenes as well.}
    \label{fig:incremental-exp}\vspace{-4mm}
\end{figure*}

In our third experiment, we used all of the objects and scenes in our dataset and similarly to the former experiments ran both of the algorithms for 200 iterations.

The learning curves for the three experiments show that the proposed Sum GP UCB approach achieves faster convergence and displays improved sample efficiency compared to the Naive BO approach. Even when providing the baseline method with a full reward evaluation, our proposed approach with only one simulation per iteration performs better. This result shows that even when simulations can be run in parallel, our proposed approach provides a performance boost. This increase in performance is due to the structure of the reward model. By modeling each observation's reward function separately and only taking the relevant object parameters as input, the model achieves better generalization by effectively capturing that parameter changes for non-present objects do not affect the reward. The benefits of the reward structure can be seen when we compare the performance across the three experiments. The benefit of the proposed approach is more noticeable in the later experiments. The larger sets of objects for these experiments result in a higher-dimensional reward modeling problem for the Naive BO. With limited generalization, more samples are required to cover the parameter space.

Looking at the selected simulations across iterations, we can see that the proposed approach results in a fairly structured exploration of the different simulations. After initially exploring each of the simulations, the robot tends to focus on subsets of simulations at a time, resulting in distinct bands in the figures. We can also see that the number of runs for each observation is not balanced, with the robot often focusing more simulations on the more challenging observations. For example, the ladle-slime simulation, which effectively only has one object whose parameters need to be determined, tends to have very few simulations across all three of the experiments. 

\subsection{Incremental Observation Set Size Experiment}

As the reward function is based on real-world observations, the reward function may change as new observations are gathered by the robot. In such cases, the reward function expands to include additional observation rewards $r(o_i,\theta_{K_i})$. By modeling each observation reward individually, our approach can easily handle an incrementally expanding reward function. The new observation automatically increases the uncertainty with its respective GP.

In this experiment, we introduce the observations in stages over the course of the full experiment. We begin by giving the robot observation 1-3 as the same as the experiment 1. After 50 iterations we expand to including observations 5-7, which introduces the sponge object as experiment 2. Finally, after a total of 100 iterations, we introduce the closed bag with observations 4 and 8. The robot thus ultimately has access to all of the observations, similar to the previous experiment 3, but the observations are added incrementally. 

The resulting learning curve is shown in Fig. \ref{fig:incremental-exp}. We illustrated the incremental learning curve with the learning curves of our method that we found in experiments 1-3. The plot thus allows us to compare the incremental performance to the performance if we had all of the corresponding observations from the start. 

The initial performance is very similar to those of Experiment 1, as one would expect. When the additional observations are given to the robot, we can see a slight overshoot in the error compared to Experiments 2 and 3. This overshoot is to be expected as the robot needs to determine the values of previously unseen parameters. Importantly though, the performances at these stages are still better than the performances seen in Experiments 2 and 3 at iteration 0. This indicates that the proposed approach is better than resetting the reward model whenever a new observation is acquired. Given the small overshoot, the incremental approach quickly reaches the same performance level as seen in Experiments 2 and 3. In other words, the robot with incremental observations quickly performs as well as if it had all of the observations from the start. The reason for this result can be seen in the plot of selected simulations. Whenever a new set of observations is provided, the algorithm automatically switches to focusing on simulations for these new observations. In some cases, it does return to the older observations. For example, after simulating the bag-cylinder multiple times, the robot switched back to the sponge-cylinder observation for a while. This result suggests that the new observation provided additional information regarding the cylinder object, which then had to be consolidated with the previous cylinder observation. The result demonstrates that the robot effectively creates a training curriculum for itself using the proposed BO strategy.


\vspace{-2pt}

\section{Conclusion}

In this paper, we proposed a Bayesian optimization approach to identifying the parameters of the objects based on observations containing different sets of interacting objects. Our approach models the rewards for individual observations using the corresponding object parameters, and it allows for partial evaluations of the reward. We showed that the resulting approach is able to identify the object parameters using significantly fewer simulation runs and scales better to larger numbers of objects and observations. 


\addtolength{\textheight}{-5cm}   




\bibliographystyle{IEEEtran}
\bibliography{root}
\end{document}